\documentclass[10pt]{article} 
\usepackage{colm2024_conference}
\usepackage{graphicx}

\usepackage{amsmath,amsfonts,bm}

\def\eqref#1{equation~\ref{#1}}

\def\1{\bm{1}}

\DeclareMathAlphabet{\mathsfit}{\encodingdefault}{\sfdefault}{m}{sl}
\SetMathAlphabet{\mathsfit}{bold}{\encodingdefault}{\sfdefault}{bx}{n}

\usepackage{url}
\usepackage{booktabs}
\usepackage{array}
\usepackage{makecell}
\usepackage{adjustbox}
\usepackage[skip=10pt]{caption}
\usepackage{multirow}
\usepackage[table]{xcolor}
\definecolor{DarkBlue}{RGB}{25, 60, 184}

\usepackage[colorlinks=true, pdfborder={0 0 0}, linkcolor=DarkBlue, citecolor=DarkBlue, urlcolor=DarkBlue]{hyperref}  

\title{Instella: Fully Open Language Models with Stellar Performance}

\author{Jiang Liu, Jialian Wu, Xiaodong Yu, Yusheng Su, Prakamya Mishra, Gowtham Ramesh, Sudhanshu Ranjan, Ximeng Sun, Ze Wang, Chaitanya Manem, Pratik Prabhanjan Brahma, Zicheng Liu, Emad Barsoum
\\[0.75em]
\bf AMD
}

\makeatletter
\@ifundefined{ps@firstpage}{\let\ps@firstpage\ps@fancy}{}
\@ifundefined{ps@undefinedpagestyle}{\let\ps@undefinedpagestyle\ps@plain}{}
\makeatother
\begin{document}

\maketitle
\begin{abstract}
Large language models (LLMs) have demonstrated remarkable performance across a wide range of tasks, yet the majority of high-performing models remain closed-source or partially open, limiting transparency and reproducibility. In this work, we introduce Instella, a family of fully open three billion parameter language models trained entirely on openly available data and codebase. Powered by AMD Instinct™ MI300X GPUs, Instella is developed through large-scale pre-training, general-purpose instruction tuning, and alignment with human preferences. Despite using substantially fewer pre-training tokens than many contemporaries, Instella achieves state-of-the-art results among fully open models and is competitive with leading open-weight models of comparable size. We further release two specialized variants: Instella-Long, capable of handling context lengths up to 128K tokens, and Instella-Math, a reasoning-focused model enhanced through supervised fine-tuning and reinforcement learning on mathematical tasks. Together, these contributions establish Instella as a transparent, performant, and versatile alternative for the community, advancing the goal of open and reproducible language modeling research.
\end{abstract}

\begin{figure}[h]
    \centering
    \includegraphics[width=\linewidth]{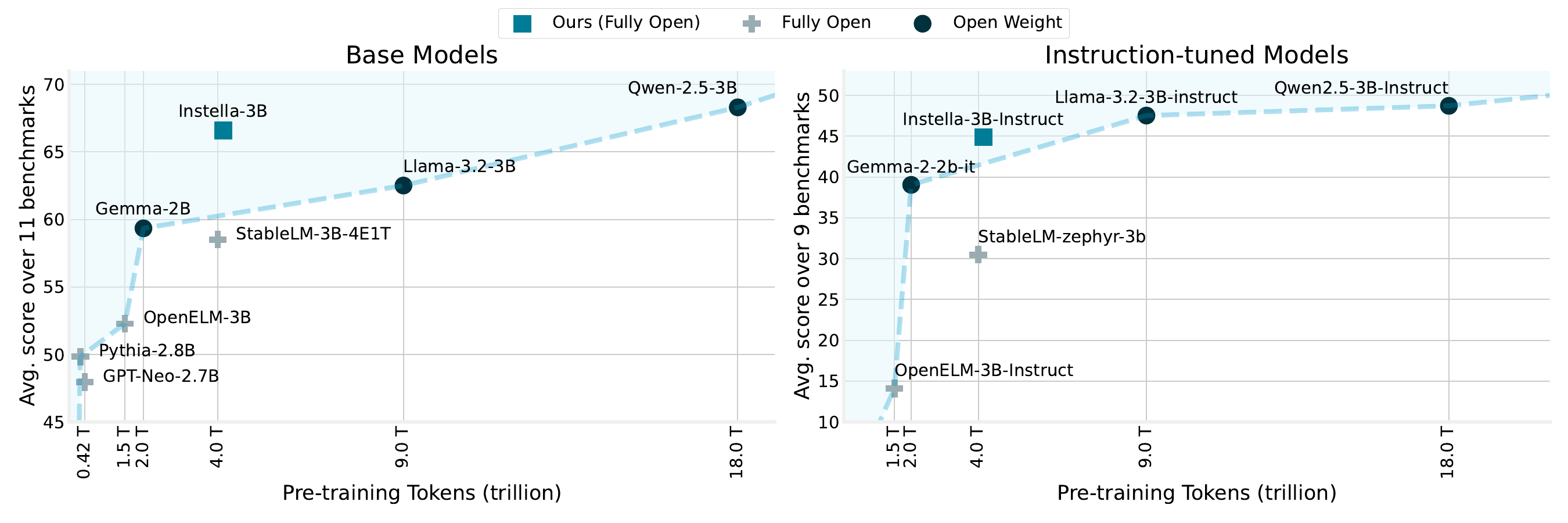}
    \caption{\textbf{Average Score versus Pre-training Tokens} for base (left) and instruction-tuned (right) models. Instella surpasses prior fully open models of comparable size and, despite being trained on substantially fewer pre-training tokens, achieves competitive performance with state-of-the-art open-weight models for both \textbf{(left)} base models (Table~\ref{tab:pretrain}) and \textbf{(right)} instruction-tuned models (Table~\ref{tab:posttrain}).}
    \label{fig:scaling}
\end{figure}

\section{Introduction}
The rapid advancement of artificial intelligence, driven in large part by large language models (LLMs)~\citep{gemini,gpt4,dubey2024llama,yang2025qwen3}, has accelerated progress toward artificial general intelligence and transformed society at large. However, much of this progress has been led by proprietary releases (e.g., GPT-4~\citep{gpt4}, Claude~\citep{claude4}, Gemini~\citep{gemini}), where training data, methods, and evaluation details remain opaque. While these models have set new state-of-the-art performance, their closed nature hinders scientific understanding, reproducibility, and equitable access.

In response, the research community has placed increasing emphasis on open-weight models, where trained parameters are released. Projects such as LLaMA-3.2-3B~\citep{dubey2024llama}, Qwen-2.5-3B~\citep{yang2024qwen2-5}, and Gemma-2-2B~\citep{team2024gemma} have demonstrated competitive capabilities in relatively compact architectures. Yet most of these remain open-weight rather than fully open: their training data, preprocessing, and training recipes are either undisclosed or proprietary. As a result, researchers cannot fully reproduce the results, audit potential data contamination, or study the effects of data and training choices at scale.

To bridge this gap, we introduce Instella, a new family of fully open 3B-parameter language models. Instella makes available not only model weights, but also the complete training pipeline, datasets, and optimization details, thereby offering full transparency. Instead of solely relying on general-purpose corpora, Instella is pretrained in two distinct stages: an initial 4T-token general-domain pre-training stage, followed by a 57B-token second-stage emphasizing reasoning-heavy domains. To further enrich this stage, we introduce an in-house synthetic dataset for mathematics, constructed by abstracting GSM8K problems into symbolic Python programs and parameterizing them to generate diverse yet solvable variants. This approach expands mathematical coverage while maintaining the correctness of synthesized data, providing a principled way to inject reasoning signals into pre-training. In addition, we leverage weight ensembling across stochastic pre-training seeds by conducting multiple second-stage runs with different random seeds and merging their weights into the final checkpoint, which further enhances model performance. Following pre-training, Instella undergoes supervised fine-tuning (SFT) on a carefully curated mixture of 2.3 million high-quality instruction-response pairs drawn from diverse domains such as mathematics, coding, commonsense reasoning, and multi-turn dialogue. This step equips the model with the ability to follow user prompts, handle complex instructions, and generalize across a wide range of task formats, and is further refined through direct preference optimization (DPO)~\citep{dpo}, aligning outputs with human expectations for helpfulness, safety, and factuality.

Building on this foundation, we extend Instella into the long-context regime with Instella-Long, capable of processing sequences up to 128K tokens. Instella-Long is trained in two stages of continued pre-training on 40B tokens, followed by long-context SFT and short-context DPO. Because of the limited availability of long-context SFT data, we synthesize long-context instruction-following examples directly from pre-training documents. Compared with other open-weight models, Instella-Long delivers competitive performance on the challenging Helmet benchmark~\citep{yen2024helmet}, while fully releasing its training details and data to ensure transparency and reproducibility.

Finally, Instella advances reasoning-centric reinforcement learning at small scale through Instella-Math. Using only 3B parameters, Instella-Math is, to our knowledge, the first fully open model of this size to apply multi-stage group relative policy optimization (GRPO)~\citep{shao2024deepseekmathpushinglimitsmathematical} entirely on open datasets. By gradually increasing rollout lengths and incorporating Olympiad-level problems from DeepScaleR~\citep{deepscaler2025}, the model demonstrates substantial improvements in mathematical and logical reasoning. Remarkably, Instella-Math performs strongly not only on benchmarks like GSM8K and OlympiadBench~\citep{he2024olympiadbench} but also on TTT-Bench~\citep{mishra2025tttbenchbenchmarkevaluatingreasoning}, highlighting that reinforcement learning can meaningfully enhance reasoning even for compact models.

Despite being trained on significantly fewer tokens compared to some leading models, Instella achieves state-of-the-art results among fully open models and rivals the performance of stronger open-weight models. To summarize, our contributions are threefold:
\begin{itemize}
    \item \textbf{Instella}. A 3B-parameter language transformer trained with a carefully staged pre-training process. Instella significantly outperforms prior fully open models of comparable size across diverse benchmarks.
    \item \textbf{Instella-Long}. A long-context variant extending sequence length to 128K tokens driven by continued pre-training and synthetic QA-based long-context instruction tuning. Instella-Long attains competitive performance on the challenging long-context benchmark Helmet.
    \item \textbf{Instella-Math}. A reasoning-centric variant fine-tuned with curated math datasets and reinforcement learning, delivering strong gains on AIME, OlympiadBench, and GSM8K while achieving the highest reported performance on the strategic reasoning benchmark TTT-Bench among fully open models.
\end{itemize}

Our work demonstrates that openness and competitiveness are not mutually exclusive. By releasing model weights, training code, data recipes, and evaluation protocols, Instella enables transparent benchmarking, reproducibility, and further research into the foundations of language modeling.
\section{Background}
\subsection{Open-Weight versus Fully-Open Large Language Models}
The release of open-weight large language models such as LLaMA~\citep{touvron2023llama, dubey2024llama} and Qwen~\citep{bai2023qwen, yang2024qwen2-5, yang2025qwen3} series has significantly broadened community access to high-performing models.
These systems are compact enough to be fine-tuned on modest hardware, enabling academic research and downstream applications.
However, most such models are not completely transparent: their pre-training datasets, training pipelines, and optimization hyperparameters remain undisclosed.
This opacity prevents reproducibility, makes data contamination difficult to audit, and constrains the ability to study scaling laws or understand how training data composition affects downstream performance.

In contrast, completely transparent models release not only weights but also data recipes, preprocessing scripts, and training code.
Notable examples include OLMo~\citep{olmo, olmo2} and SmolLM~\citep{allal2025smollm2smolgoesbig}, which provide comprehensive training pipelines and fully specified data mixtures.
These initiatives enable researchers to systematically investigate questions such as how data diversity affects generalization, how alignment methods interact with model size, and how pre-training choices influence reasoning capabilities.
However, prior fully open 3B models still underperform compared to state-of-the-art open-weight systems by a considerable margin on challenging benchmarks such as GSM8K~\citep{gsm}, BBH~\citep{bbh}, and MMLU~\citep{mmlu}, motivating further work to bridge the gap between transparency and competitiveness. Instella addresses this gap by offering a fully open 3B-parameter model family with state-of-the-art results.
We release not only weights but also training data recipes, preprocessing scripts, optimization settings, and evaluation pipelines, providing a truly reproducible foundation for scientific study.

\subsection{Long-context Language Models}
Many real-world applications demand reasoning over inputs significantly longer than the typical 2K–8K context windows used in base large language models.
Tasks such as legal document analysis, multi-chapter summarization, and retrieval-augmented generation require context lengths exceeding 100K tokens.
Recent advances including efficient attention mechanisms~\citep{dao2023flashattention2,jacobs2023deepspeed,liu2023ring}, rotary position embedding (RoPE) scaling~\citep{ropescale, dynamicNTK, ding2024longrope}, and specialized training strategies for long sequences~\citep{gao2024train} have enabled models to process extended sequences. Despite these developments, few transparent models provide both long-context support and strong performance. On the other hand, open-weight models such as Qwen2.5-1M~\citep{qwen2.5-1m} offer extended context windows, but their training data remain proprietary, limiting reproducibility. Instella-Long contributes to this space by transparently extending the context length to 128K tokens through continued pre-training and post-training on the long-context data we release publicly. It achieves competitive results on the long-context benchmarks while establishing a transparent, reproducible long-context baseline.

\subsection{Large Reasoning Models}
The ability to perform multi-step reasoning represents a central goal for large language model development.
Benchmarks such as MMLU, BBH, GSM8K, MATH~\citep{hendrycksmath2021} and AIME~\citep{AIME} measure a model's capacity to perform structured, compositional thinking beyond surface-level pattern matching.
Recent research demonstrates that high-quality reasoning data and post-training techniques such as reinforcement learning can dramatically improve performance.
Models like DeepSeek-R1~\citep{deepseekai2025deepseekr1incentivizingreasoningcapability} and DeepSeek-Math~\citep{shao2024deepseekmathpushinglimitsmathematical} show that incorporating step-by-step solutions and applying alignment methods like group relative policy optimization (GRPO)~\citep{shao2024deepseekmathpushinglimitsmathematical} can lead to substantial gains in reasoning capabilities.

However, most reasoning-focused models remain only partially open: either the reasoning datasets are proprietary, the reinforcement learning recipes are undisclosed, or the resulting models are released without reproducible training pipelines.
This lack of transparency hinders systematic study of reasoning capabilities and prevents independent validation of methodological claims.

Instella-Math addresses this limitation by providing the first fully open 3B-parameter model trained with multi-stage reinforcement learning entirely on open data.
We release not only the model weights but also the reasoning datasets and training configurations, enabling reproducible research into reasoning emergence and reinforcement learning training for small-scale models.

\section{Instella}
\subsection{Model Architecture}
The Instella models are text-only, autoregressive transformer-based language models~\citep{attention} with 3 billion parameters. Architecture-wise, Instella consists of 36 decoder layers, each having 32 attention heads with a hidden dimension of 2,560 and an intermediate dimension of 6,912. We use standard multi-head attention~\citep{attention}. For layer normalization, we employ RMSNorm~\citep{rmsnorm}, which has been shown to provide better training stability and convergence properties compared to standard LayerNorm~\citep{layernorm}, particularly for large-scale language models~\citep{takase2023spike, touvron2023llama, olmoe}.

In addition, we apply QK-Norm~\citep{qknorm, olmoe,naseer2021intriguing}, where layer normalization is injected after the query and key projections within each attention head. QK-Norm normalizes the query and key vectors before computing attention scores, helping to maintain more balanced attention distributions throughout training. It has been shown to be effective in improving training stability by preventing attention weights from becoming overly extreme, which can lead to gradient instability and poor convergence. 

Our model uses a standard causal attention mask. The feed-forward network within each transformer layer follows the standard architecture with SwiGLU activation function, which has demonstrated superior performance compared to ReLU-based activations in recent language models. We also employ rotary position embeddings (RoPE)~\citep{su2024roformer} to encode positional information, which provides better extrapolation to longer sequences compared to absolute positional embeddings.

The key hyperparameters of Instella-3B architecture are shown in Table~\ref{tab:model_arch}. We use the OLMo tokenizer~\citep{olmo} with a vocabulary size of 50,304 tokens. This vocabulary size strikes a balance between computational efficiency and representation capacity, allowing the model to handle diverse text while maintaining reasonable embedding and output layer sizes.

\begin{table}[h]
    \centering
    \caption{Key hyper-parameters of Instella-3B architecture.}

    \begin{tabular}{ccccccc}
    \toprule
    Number of          & Hidden  &  Intermediate   & Number of       & Number of & Sequence & Vocabulary \\
    transformer layers & dimension & dimension & attention heads &    KV heads                & length   & size \\
    \toprule
    36 & 2560 & 6912 &  32 & 32 & 4096 & 50,304 \\
    \bottomrule
    \end{tabular}

    \label{tab:model_arch}
\end{table}

\subsection{Training Setup}

Our training pipeline is based on the open-sourced OLMo codebase, adapted, and optimized for our hardware and model architecture. For pre-training we use a total of 128 Instinct MI300X GPUs distributed across 16 nodes. During both pre-training and post-training, we utilize FlashAttention 2~\citep{dao2023flashattention2}, Torch Compile, and bfloat16 mixed-precision training to reduce memory usage and speed up training. To balance inter-node memory efficiency and intra-node communication overhead within our cluster, we employ fully sharded data parallelism (FSDP) with hybrid sharding, with model parameters, gradients, and optimizer states sharded within a node and replicated across the nodes. 



\subsection{Pre-training}
\begin{figure}[h]
    \centering
    \includegraphics[width=0.8\linewidth]{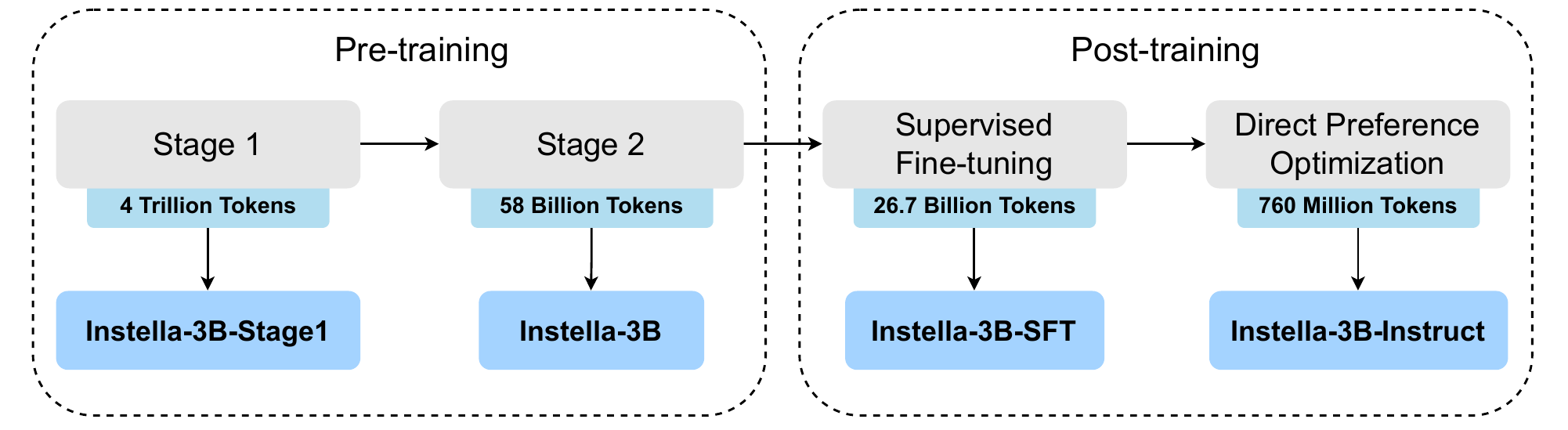}
    \caption{Instella-3B model training pipeline.}
    \label{fig:pipeline}
\end{figure}

We pre-train the model using two stages with a sequence length of 4,096 tokens and a global batch size of 1,024. The Instella 3B pretraining pipeline is shown in Fig.~\ref{fig:pipeline}. In the first pre-training stage, we train the model from scratch on 4.07 trillion tokens sourced from OLMoE-mix-0924~\citep{olmoe}, which is a diverse mix of two high-quality datasets DCLM-baseline~\citep{dclm} and Dolma 1.7~\citep{dolma} covering domains like coding, academics, mathematics, and general world knowledge from web crawl. This extensive first stage pre-training established a foundational understanding of general language in our Instella model. We use the cosine decay learning rate schedule with a maximum learning rate of $4\times 10^{-4}$ and set the global batch size to 1024.

For our final pre-trained checkpoint, Instella-3B, we conduct a second stage pre-training on top of the first-stage Instella-3B-Stage1 model to further enhance its capabilities on MMLU~\citep{mmlu}, BBH~\citep{bbh}, and GSM8K~\citep{gsm}. The model is trained three times with different random seeds, and the resulting weights are ensembled to obtain the final checkpoint. Specifically, the second-stage training uses 58 billion tokens sourced from diverse and high-quality datasets, including Dolmino-Mix-1124~\citep{olmo2}, SmolLM-Corpus (python-edu)~\citep{smollmcorpus}, Deepmind Mathematics~\citep{dmmath}, and conversational datasets such as Tülu-3-SFT-Mixture~\citep{tulu3}, OpenHermes-2.5~\citep{OpenHermes}, WebInstructSub~\citep{webinstruct}, Code-Feedback~\citep{opencodeinterpreter}, and Ultrachat 200k~\citep{ultrachat}.  We use the linear decay learning rate schedule with a maximum learning rate of $4\times 10^{-5}$ and set the global batch size to 1024. 

In addition to the publicly available datasets, 28.5 million tokens in the second-stage pre-training data mixture are derived from our in-house synthetic dataset focused on mathematical problems. This dataset is generated using the training set of GSM8k dataset, where we first use Qwen2.5-72B-Instruct~\citep{yang2024qwen2-5} to 1) abstract numerical values as function parameters and generate a python program to solve the math question, 2) identify and replace numerical values in the existing question with alternative values that are still answerable with the same python program solution as the original question. Next, by assigning different new values to these python parameters and using the abstract solution program to compute the corresponding answers, we expand our synthetic dataset with new and reliable question-answer pairs~\citep{yu2024reasonagain}. 

\subsection{Post-training}
We first perform supervised finetuning (SFT) to enable the pre-trained model to follow instructions and respond effectively to user queries. We train for three epochs on 2.3 millions of high-quality instruction–response pairs, resulting in Instella-3B-SFT. During this phase, we utilize datasets spanning a broad spectrum of tasks and domains to ensure that the model generalizes across diverse instruction types. The mixture is selectively sourced from SmolTalk~\citep{allal2025smollm2smolgoesbig}, OpenMathInstruct-2~\citep{toshniwal2024openmathinstruct}, Tülu-3 Instruction Following~\citep{tulu3}, MMLU auxiliary train set~\citep{mmlu}, and o1-journey~\citep{o1journey}. We use the linear decay learning rate schedule with a maximum learning rate of $1\times 10^{-5}$ and set the global batch size to 128. 

In the final training stage, we align Instella-3B-SFT with human preferences to ensure its outputs are helpful, accurate, and safe. Building on Instella-3B-SFT, Instella-3B-Instruct is trained with direct preference optimization (DPO)~\citep{dpo} on 0.76 billion tokens from the OLMo 2 1124 7B Preference Mix~\citep{olmo2}. This alignment step tailors the model’s responses to better reflect human values and expectations, thereby improving the quality and reliability of its outputs. We use the linear decay learning rate schedule with a maximum learning rate of $5\times 10^{-7}$ and set the global batch size to 128.

\section{Instella-Long}
\begin{figure}[h]
    \centering
    \includegraphics[width=0.8\linewidth]{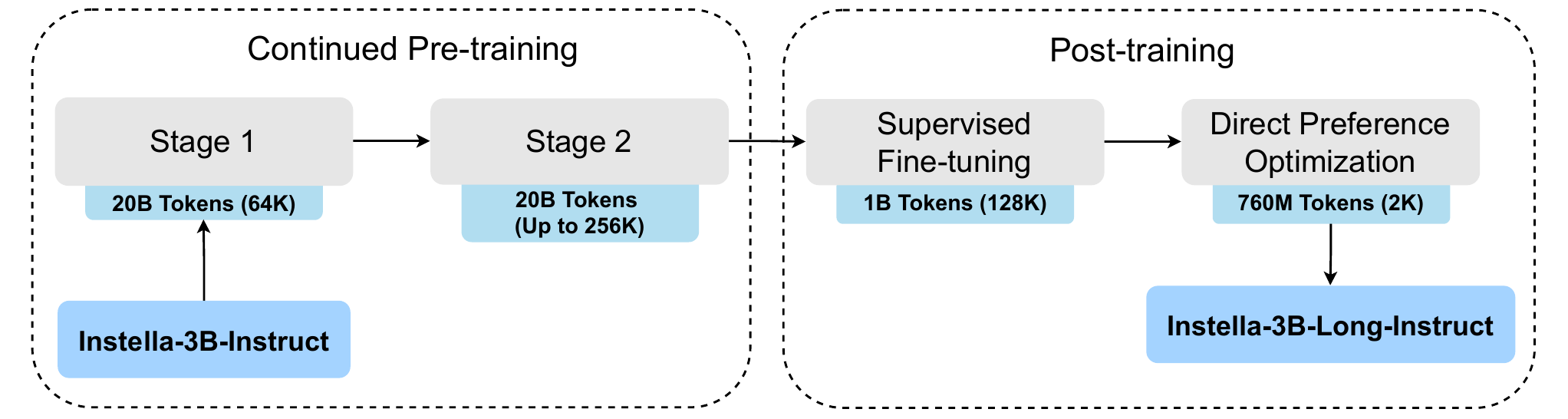}
    \caption{Instella-Long model training pipeline.}
    \label{fig:pipeline-long}
\end{figure}

In this section, we introduce the long-context model of Instella, namely, Instella-3B-Long-Instruct, supporting 128K context length. To extend the context length, we continually train the model from Instella-3B-Instruct through: 1. continued pre-training, 2. supervised finetuning (SFT), and 3. direct preference optimization (DPO), as shown in Fig. \ref{fig:pipeline-long}. We detail the training method and data in the following subsections.

\subsection{Continued Pre-training}
The long context training is initialized from the short-context checkpoint, Instella-3B-Instruct, which has a context length of 4K. We conduct a two-stage continued pre-training to gradually increase the context length. \textbf{Stage 1}: We extend the context length from 4K to 64K and train the model using 20B tokens. The batch size is 4M tokens and the training steps are 5,000. We follow the RoPE scaling law~\citep{ropescale} to increase the base frequency of RoPE from 10,000 to 514,640. We also experiment with alternative RoPE scaling methods~\citep{dynamicNTK,gao2024train} and observe only minor differences in performance.
\textbf{Stage 2}: As indicated by~\citep{gao2024train}, it is beneficial to train the model with the data whose context length is longer than the target context length. In this stage, we train the model on 20B tokens with a maximum context length of 256K - twice our target context length of 128K. Following the RoPE scaling law, we further increase the RoPE base frequency to 3,691,950. The batch size is 8M tokens and the training steps are 2,500. For both stages, we use the linear decay learning rate schedule and the maximum learning rate is $2\times10^{-5}$.

\begin{table}[htbp]
\caption{Long-context continued pre-training data by source and portion. Each stage consists of 20 billion tokens in total.}
\label{tab:long_pretrain_data}
\small
\centering
\begin{tabular}{cccc}
\toprule
\textbf{Training Stage} & \textbf{64K Long Data} & \textbf{256K Long Data} & \textbf{Short Data} \\
\midrule
Stage 1 & 
\makecell{Code repos (30\%)\\Books (30\%)\\Textbooks (3\%)} & 
-- & 
\makecell{FineWeb-Edu (10\%)\\FineWeb (10\%)\\Wikipedia (5\%)\\OpenWebMath (5\%)\\StackExchange (4\%)\\ArXiv (3\%)} \\
\hline
Stage 2 & 
\makecell{Code repos (10\%)\\Books (15\%)} & 
\makecell{Code repos (20\%)\\Books (15\%)\\Textbooks (2\%)} & 
\makecell{FineWeb-Edu (10\%)\\FineWeb (10\%)\\Wikipedia (5\%)\\OpenWebMath (5\%)\\StackExchange (4\%)\\ArXiv (4\%)} \\
\bottomrule
\end{tabular}
\end{table}
The continued pre-training data originates from the data mixture created by Prolong~\citep{gao2024train}. We use the raw text data curated by Prolong and process the data through tokenization, filtering, and packing. In each stage of the continued pre-training, we train on a 20B-token mixture of short- and long-context data with an approximate ratio of 4 to 6. The detailed data sources and portion are listed in Table~\ref{tab:long_pretrain_data}. Let $L$ be the maximum context length of the training stage. We pack both short- and long-context data into $L$-length sequences for training. For short-context data, we randomly select multiple documents and concatenate them into an $L$-length sequence. The extra texts beyond $L$ in the last document are discarded. For long-context data, we filter out the documents that are shorter than $L$. We observe that the raw text data has some super long documents ($>>L$). For these documents, we randomly sample a few segments from them to avoid producing an excessive number of training examples from a single document.  We mix 64K data into the long-context data in the second stage for improving training throughput, where we pack four different 64K documents into a 256K sequence. During data processing, we ensure that the documents used in the first and second stages are mutually exclusive. In training, we apply document masking so that different documents within the same sequence cannot attend to each other.

\subsection{Post-training}

After continued training on the long-context pre-training data, we perform supervised finetuning on a 1B-token mixture of short- and long-context instruction data. We use a batch size of 4M tokens and train for 250 steps. A linear decay learning rate schedule is employed, with a maximum learning rate of $4\times10^{-5}$. For the SFT data, we pack multiple samples into a 256K sequence with document masking applied during training. Padding tokens are added in order to reach exactly 256K tokens.

Similar to the continued pre-training, we train the model on a mixture of short- and long-context instructions data with a ratio of 4 to 6. For short-context instruction data, we use publicly available instruction-tuning datasets, some of which are also used in the post-training of Instella-3B-Instruct. Specifically, we use Ultrachat 200K~\citep{ultrachat}, OpenMathinstruct-2~\citep{toshniwal2024openmathinstruct}, Tülu-3 Instruction Following~\citep{tulu3}, and MMLU auxiliary train set~\citep{mmlu}. 

Due to the lack of long-context SFT data, we construct a long-context instruction-following dataset where the context length is controlled to be between 8K and 128K tokens. Specifically, we make use of the long-context documents of Books from our continued pre-training data corpus. We use the documents that have at least 8K tokens and truncate the document to 128K tokens if it is over 128K. Then, we use Qwen2.5-14B-Instruct-1M~\citep{qwen2.5-1m} as a teacher model to synthetically generate a question and an answer for the document. To speed up this process, we randomly choose a subpart of the document for the QA generation instead of using the whole document. The length of the subpart is randomly set to be between 2K and 8K tokens. We use NLTK~\citet{bird-loper-2004-nltk} sentence tokenizer to divide documents into sentences to make sure that the selected subpart has complete sentences. The generated question and answer are appended to the end of the long document, serving as a complete single-round instruction-following data sample. Furthermore, we generate long-context instruction data from short-context documents, thereby enhancing dataset diversity with a broader range of sources. We use ArXiv from our continued pre-training corpus and the DCLM subset from Dolmino-Mix-1124~\citep{olmo2}. We first generate QA for each short-context document following the same pipeline aforementioned. Next, we iteratively concatenate different short-context documents into a long sequence until it reaches 128K tokens. Since we do not truncate the last document, the concatenated sequence may exceed 128K tokens. Lastly, we randomly choose one QA corresponding to one of the short-context documents and append it to the end of the concatenated sequence. Contrary to the findings by~\citep{gao2024train}, we observe that our synthetic long-context instruction data notably improves performance on long-context tasks. The final SFT data mixture is shown in Table~\ref{tab:long_sft_data}.

\begin{table}[htbp]
\caption{Long-context supervised finetuning data by source and portion, totaling 1 billion tokens.}
\label{tab:long_sft_data}
\small
\centering
\begin{tabular}{c|c}
\toprule
\textbf{Short Data} & \textbf{Long Data} \\
\midrule
 \makecell{Ultrachat 200K (25\%), OpenMathinstruct-2 (10\%),\\MMLU auxiliary train set (3\%),\\ Tülu-3 Instruction Following (2\%)} & 
\makecell{Books (44\%), DCLM (10\%), ArXiv (6\%)} \\
\bottomrule
\end{tabular}
\end{table}

In the final training stage, we perform human preference alignment using DPO~\citep{dpo}, employing the same training setting and dataset as Instella-3B-Instruct. Different from the previous long-context training stages, this DPO stage is trained on short-context data only with a maximum context length of 2K. Consistent with the findings of other open-weights models, we observe that applying DPO solely on short-context data continues to improve on long-context tasks.

\subsection{Implementation Details}
\textbf{Sequence Parallelism.} We implement sequence parallelism based on Deepspeed Ulysses~\citep{jacobs2023deepspeed}, which distributes the attention heads across GPUs during attention computation. Compared to Ring-Attention~\citep{liu2023ring}, this approach is more communication-efficient. For the second continued pre-training stage and SFT, we employ four GPUs as a sequence parallelism group to handle the long input sequences. Sequence parallelism is not used in other stages, as the memory requirements fit within a single GPU.

\textbf{Document Masking and Data Batching.} We apply document masking during the continued pre-training and SFT, as each input sequence may contain multiple documents. Document masking is achieved through variable-length FlashAttention~\citep{dao2023flashattention}, which computes attention within each individual document rather than across the entire sequence. This design can also improve training throughput when combined with sorted data batching. Following Prolong~\citep{gao2024train}, we sort microbatches at each training step by the sum of document lengths in the sequence. With gradient accumulation, later microbatches benefit from faster processing when they consist of shorter documents.

\section{Instella-Math}
\begin{figure}[h]
    \centering
    \includegraphics[width=0.8\linewidth]{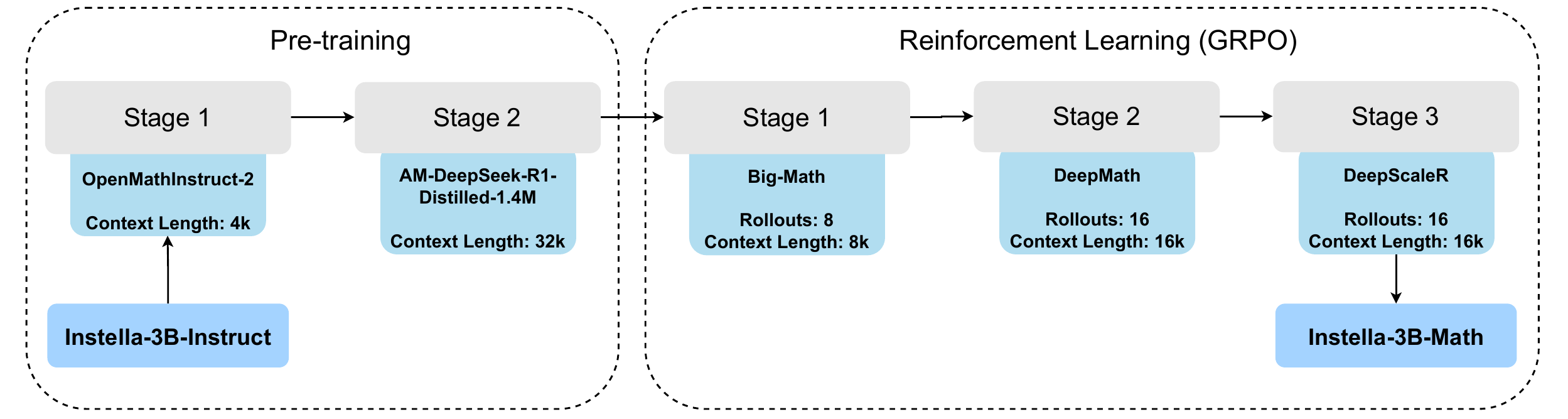}
    \caption{Instella-Math model training pipeline.}
    \label{fig:math_pipeline}
\end{figure}

In this section, we introduce Instella-Math, a reasoning-centric language model trained with long chain-of-thought reinforcement learning. To enhance the model’s mathematical and logical reasoning capabilities, we continually train Instella-3B-Instruct through two stages of supervised finetuning and three stages of reinforcement learning, as shown in Figure \ref{fig:math_pipeline}. We detail the training procedure and datasets below.

\subsection{Supervised Finetuning}

As a cold start, we perform a two-stage supervised finetuning process to enhance the reasoning capabilities of Instella-3B-Instruct:

\textbf{Stage 1: Instruction Tuning with OpenMathInstruct-2 for Mathematical Coverage}. In the first SFT stage, we begin with instruction tuning, following instructions or prompts properly, especially in a question-answer or problem-solution format. Using the OpenMathInstruct-2 dataset~\citep{toshniwal2024openmathinstruct}, which consists of 14 million problem-solution pairs generated from the GSM8K~\citep{gsm} and MATH~\citep{hendrycksmath2021} training sets, the model is trained to solve mathematical questions covering a diverse range of topics from arithmetic and algebra to probability and calculus.

\textbf{Stage 2: Deep Reasoning with Long-Context Training on AM-DeepSeek-R1-Distilled}. In the second SFT stage, we further improve the model’s reasoning capability by training on AM-DeepSeek-R1-Distilled-1.4M~\citep{zhao202514millionopensourcedistilled}, which is a large-scale general reasoning dataset containing high-quality and challenging problems. In this stage, we increase the context length of the model from 4K to 32K to allow the model to learn from the long chain-of-thought responses distilled from large reasoning models such as DeepSeek-R1 \citep{deepseekai2025deepseekr1incentivizingreasoningcapability}.

\subsection{Reinforcement Learning}

Following supervised finetuning, we apply three stages of reinforcement learning using the group relative policy optimization (GRPO) algorithm \citep{shao2024deepseekmathpushinglimitsmathematical} to further strengthen the model's mathematical reasoning abilities. Training is orchestrated with verl~\citep{sheng2024hybridflow} and vLLM~\citep{kwon2023efficient} for efficient rollout collection, reward scoring, and policy updates.

\textbf{Stage 1: GRPO on Big-Math-RL-Verified (8 Rollouts @ 8K Tokens)}. In the first stage of reinforcement learning, we apply the GRPO algorithm to train the model on Big-Math-RL-Verified \citep{albalak2025bigmathlargescalehighqualitymath}, a collection of curated, complex, multi-step math problems. We generate 8 rollouts per prompt, each with up to 8K output tokens, to explore diverse reasoning trajectories. The model is trained for 1,200 GRPO steps using rule-based reward signals provided by Prime-RL~\citep{cui2025process}, which incentivize correctness and well-structured outputs.

\textbf{Stage 2: GRPO on DeepMath (16 Rollouts @ 16K Tokens)}. To push the limits of long-form reasoning, we conduct a second GRPO stage on DeepMath~\citep{deepmath} using 16 rollouts per prompt with up to 16K output tokens. This stage is designed to maximize the model’s capacity for deep mathematical reasoning, enabling it to solve problems that require extended derivations, multiple nested logical steps, or structured proof-like outputs. In this stage, the model is trained for 600 GRPO steps.

\textbf{Stage 3: GRPO on DeepScaleR (16 Rollouts @ 16K Tokens)}. In the final GRPO stage, we finetune the model on DeepScaleR~\citep{deepscaler2025}, which includes original Olympiad math problems (e.g., AIME and AMC). Similar to Stage 2, this training uses 16 rollouts and a 16K token limit. We run 740 GRPO steps in this phase to improve performance on competition-style reasoning tasks.

\section{Evaluation}


\subsection{Base Model}
\begin{table}[t]
\centering
\setlength{\tabcolsep}{4pt} 
\small
\caption{\textbf{Base model performance.}}
\begin{tabular}{l|cccccccccccc}
\toprule
\textbf{Models} & \textbf{ARC-C} & \textbf{ARC-E} & \textbf{BoolQ} & \textbf{HS.} & \textbf{PiQA} & \textbf{SciQ} & \textbf{WG.} & \textbf{OBQA} & \textbf{MMLU} & \textbf{BBH} & \textbf{GSM8K} & \textbf{Avg.} \\
\midrule
\multicolumn{13}{c}{\textit{Open Weight Models}} \\
\midrule
Gemma2-2B & 39.5 & 59.3 & 74.5 & 70.5 & 76.4 & {96.6} & 69.8 & 44.8 & 53.3 & 40.8 & 27.4 & 59.3 \\
Llama-3.2-3B & 47.2 & 64.9 & 74.8 & 73.1 & 75.9 & 95.3 & 70.3 & 51.2 & 57.8 & {47.0} & 30.1 & 62.5 \\
Qwen2.5-3B & 51.5 & 67.2 & {79.1} & 72.1 & 77.4 & 95.5 & 69.3 & {51.4} & {67.2} & {56.7} & {63.8} & {68.3} \\
\midrule
\multicolumn{13}{c}{\textit{Fully Open Models}} \\
\midrule
Pythia-2.8B & 40.5 & 60.7 & 64.8 & 60.1 & 72.5 & 89.7 & 60.8 & 42.6 & 26.1 & 27.7 & 2.7 & 49.8 \\
GPTNeo-2.7B & 38.5 & 54.6 & 62.7 & 55.2 & 70.8 & 88.0 & 58.3 & 40.8 & 27.8 & 27.3 & 3.7 & 48.0 \\
OpenELM-3B & 37.5 & 58.4 & 68.6 & 71.7 & 75.6 & 92.5 & 65.4 & 46.4 & 26.7 & 29.4 & 3.0 & 52.3 \\
StableLM-3B & 44.8 & 67.0 & 75.4 & {74.2} & {78.4} & 93.4 & 68.4 & 48.6 & 45.2 & 37.3 & 10.8 & 58.5 \\
\textbf{Instella-3B-Stage1} & {53.9} & {73.2} & {78.7} & {74.2} & 77.5 & 94.9 & {71.2} & {51.4} & 54.7 & 34.3 & 10.8 & 61.3 \\
\textbf{Instella-3B} & {52.8} & {70.5} & 76.5 & {75.0} & {77.8} & {96.4} & {73.1} & {52.4} & {58.3} & 39.7 & {59.8} & {66.6} \\
\bottomrule
\end{tabular}
\label{tab:pretrain}
\end{table}

\begin{table}[t]
\centering
\setlength{\tabcolsep}{5pt} 
\small
\caption{\textbf{Instella 3B base model performance.} We report the model performance after stage 1 and stage 2 pretraining. For stage 2, we run the training for three times with different random seeds and merge model weights to obtain the final stage 2 model.}
\begin{tabular}{l|cccccccccccc}
\toprule
\textbf{Models} & \textbf{ARC-C} & \textbf{ARC-E} & \textbf{BoolQ} & \textbf{HS.} & \textbf{PiQA} & \textbf{SciQ} & \textbf{WG.} & \textbf{OBQA} & \textbf{MMLU} & \textbf{BBH} & \textbf{GSM8K} & \textbf{Avg.} \\
\midrule
\textbf{Stage1} & {53.9} & {73.2} & {78.7} & {74.2} & 77.5 & 94.9 & {71.2} & 51.4 & 54.7 & 34.3 & 10.8 & 61.3 \\
\midrule
\textbf{Stage2-seed1} & 51.2 & 68.8 & 76.2 & 73.8 & 77.3 & {96.6} & 72.1 & {52.0} & 57.7 & 38.5 & 56.1 & 65.5 \\
\textbf{Stage2-seed2} & 50.8 & 68.4 & {77.8} & 74.3 & 77.2 & {96.6} & 71.8 & 51.4 & {58.2} & 38.5 & {58.8} & {65.8} \\
\textbf{Stage2-seed3} & 49.8 & 68.8 & 73.5 & {75.6} & 77.2 & {96.7} & {72.8} & {52.0} & 58.0 & {38.6} & 58.3 & 65.6 \\
\midrule
\textbf{Stage2} & {52.8} & {70.5} & {76.5} & {75.0} & {77.8} & 96.4 & {73.1} & {52.4} & {58.3} & {39.7} & {59.8} & {66.6} \\
\bottomrule
\end{tabular}
\label{tab:pretrain2}
\end{table}

 We evaluate the pre-trained base models on ARC-Challenge (ARC-C)~\citep{clark2018thinksolvedquestionanswering}, ARC-Easy
 (ARC-E)~\citep{clark2018thinksolvedquestionanswering}, BoolQ~\citep{clark-etal-2019-boolq}, HellaSwag (HS)~\citep{zellers-etal-2019-hellaswag}, PiQA~\citep{bisk2019piqareasoningphysicalcommonsense}, SciQ~\citep{welbl2017crowdsourcingmultiplechoicescience}, WinoGrande (WG)~\citep{sakaguchi2019winograndeadversarialwinogradschema}, OpenBookQA (OBQA)~\citep{mihaylov2018suitarmorconductelectricity}, BBH~\citep{suzgun2022challengingbigbenchtaskschainofthought}, MMLU~\citep{hendrycks2021measuringmassivemultitasklanguage}, and GSM8k~\citep{cobbe2021trainingverifierssolvemath}. All the benchmarks use a zero-shot evaluation setting, except BBH, MMLU, and GSM8k, which are evaluated using 3-shot, 5-shot, and 8-shot prompting, respectively.
 
As shown in Table ~\ref{tab:pretrain}, both Instella-3B-Stage1 and Instella-3B models outperform all the other fully open models over all the benchmarks individually (except PIQA). Our final pre-trained checkpoint Instella-3B outperforms the prior top performant fully open pre-trained models by a lead of 8.1\% on average, with significant improvements in ARC Challenge (+8\%), ARC Easy (+3.5\%), Winnograde (+4.7\%), OpenBookQA (+3.9\%), MMLU (+13.1\%) and GSM8K (+49\%).

Second stage pre-training elevates the overall average performance relative to stage-1 by 5.3\%, substantially narrowing the performance gap between Instella-3B model and the prior open weight models, and outperforming Llama-3.2-3B by 4.1\% on average (+5.7\% ARC-Challenge, +5.6\% ARC-Easy, and +29.7\% GSM8k), Gemma-2-2B by 7.3\% on average (+13.4\% ARC-Challenge, +11.2\% ARC-Easy, +4.5\% HellaSwag, +7.6\% OpenBookQA, +5.0\% MMLU, and +32.5\% GSM8k), and is competitive with Qwen-2.5-3B on the majority of the benchmarks. As shown in Table \ref{tab:pretrain2}, the Instella-3B checkpoint, obtained by merging the weights of three independently trained models with different random seeds during second stage pretraining, achieves an average performance of 66.6\%, surpassing all individual seed runs.

The multi-stage pre-training with diverse and high-quality data mixture significantly enhances Instella-3B’s capabilities, establishing it as a competitive and open alternative in the landscape of comparable size language models.
\subsection{Instruction-tuned Model}
\begin{table}[t]
\centering
\setlength{\tabcolsep}{5pt} 
\caption{\textbf{Instruction-tuned model performance.}}
\begin{tabular}{l|cccccccccc}
\toprule
\textbf{Models} & \textbf{MMLU} & \textbf{TQA} & \textbf{BBH} & \textbf{GPQA} & \textbf{GSM8K} & \textbf{MATH} & \textbf{IFEval} & \textbf{AE 2} & \textbf{MT} & \textbf{Avg.} \\
\midrule
\multicolumn{11}{c}{\textit{Open Weight Models}} \\
\midrule
Gemma-2-2B-Instruct & 58.4 & {55.8} & 43.0 & 25.2 & 53.5 & 22.5 & 55.6 & {29.4} & {8.1} & 39.0 \\
Llama-3.2-3B-Instruct & {61.5} & 50.2 & {61.5} & {29.7} & {77.0} & {46.0} & {75.4} & 19.3 & 7.1 & {47.5} \\
Qwen-2.5-3B-Instruct & {66.9} & {57.2} & {57.3} & 28.1 & {76.0} & {60.4} & 62.5 & {22.1} & {8.0} & {48.7} \\
\midrule
\multicolumn{11}{c}{\textit{Fully Open Models}} \\
\midrule
StableLM-zephyr-3B & 45.1 & 47.9 & 39.3 & 25.7 & 58.4 & 10.4 & 34.2 & 7.5 & 6.0 & 30.5 \\
OpenELM-3B-Instruct & 27.4 & 38.1 & 24.2 & 18.1 & 1.6 & 0.4 & 16.1 & 0.2 & 1.0 & 14.1 \\
\textbf{Instella-3B-SFT} & 58.8 & 52.5 & 46.0 & 28.1 & 71.7 & 40.5 & 66.2 & 7.6 & 7.1 & 42.1 \\
\textbf{Instella-3B-Instruct} & 58.9 & 55.5 & 46.8 & {30.1} & 73.9 & 42.5 & {71.4} & 17.6 & 7.2 & 44.9 \\
\bottomrule
\end{tabular}
\label{tab:posttrain}
\end{table}

The instruction-tuned models are evaluated on MMLU~\citep{hendrycks2021measuringmassivemultitasklanguage}, TruthfulQA (TQA)~\citep{lin2022truthfulqameasuringmodelsmimic}, BBH~\citep{suzgun2022challengingbigbenchtaskschainofthought}, GPQA~\citep{rein2023gpqagraduatelevelgoogleproofqa}, GSM8K~\citep{cobbe2021trainingverifierssolvemath}, Minerva Math~\citep{lewkowycz2022solving} (MATH), IFEval~\citep{zhou2023instructionfollowingevaluationlargelanguage}, Alpaca Eval V2 (AE2)~\citep{dubois2025lengthcontrolledalpacaevalsimpleway}, and MT-Bench (MT)~\citep{zheng2023judgingllmasajudgemtbenchchatbot}. Here, GPQA, Minerva Math, IFEval, and Alpaca V2 use a zero-shot evaluation setting, whereas MMLU, TQA, BBH, and GSM8k use few-shot prompting using 5-shots, 6-shots, 3-shots, and 8-shots, respectively.

Instella-3B-Instruct model consistently outperforms other fully open models across all evaluated benchmarks with a significant average score lead of 14.37\% with respect to the next top performing fully open instruction-tuned models (Table \ref{tab:posttrain}). With substantial margins across all the chat benchmarks (+13\% MMLU, +7.57\% TruthfulQA, +7.43\% BBH, +4.46\% GPQA, +37.15\% IFEval, +10.08\% Alpaca 2, and +1.2\% MT-Bench).

Instella-3B-Instruct narrows the performance gap with leading open-weight models. Instella-3B-Instruct performs on par with or slightly surpasses existing state-of-the-art open weight instruction-tuned models such as Llama-3.2-3B-Instruct (+5.24\% TruthfulQA, +0.45\% GPQA, and +0.1\% MT-Bench), and Qwen2.5-3B-Instruct (+2.01\% GPQA and +8.87\% IFEval), while significantly outperforming Gemma-2-2B-Instruct with an average score lead of +5.83\% (+0.55\% MMLU, +3.79\% BBH, +4.91\% GPQA, +20.47\% GSM8k, +19.98\% Minerva MATH, and +15.17\% IFEval).

Overall, Instella-3B-Instruct excels in instruction following tasks and multi-turn QA tasks like TruthfulQA, GPQA, IFEval and MT-Bench, while being highly competitive compared to existing state-of-the-art open weight models on other knowledge recall and math benchmarks, while being trained on significantly fewer training tokens.

\subsection{Instella-Long}
\begin{table*}[htbp]
\centering
\setlength{\tabcolsep}{4pt}
\caption{\textbf{Long-context evaluation on the Helmet benchmark.} NQ: Natural Question. Inf: InfiniteBench. NarrQA: NarrativeQA. The NIAH-MV task and RAG task (NQ, TriviaQA, and HotpotQA) are evaluated at five context lengths: 8K, 16K, 32K, 64K, and 128K, and the number is reported by averaging across the five context lengths. The InfQA, InfMC, and NarrQA are evaluated at 128K context length.}
\label{tab:long_main_results}
\begin{tabular}{lccccccccc}
\toprule
\textbf{Models} & \textbf{NQ} & \textbf{TriviaQA} & \textbf{HotpotQA} & \textbf{InfQA} & \textbf{InfMC} & \textbf{NarrQA} & \textbf{NIAH-MV} & \textbf{Avg.} \\
\midrule
\multicolumn{9}{c}{\textit{Open Weight Models}} \\
\midrule
Llama-3.2-3B-Instruct   & 51.8 & 86.2 & 56.4 & 38.7 & 56.0 & 26.0 & 99.2 & {59.2} \\
Phi-3.5-Mini-Instruct          & 41.2 & 78.6 & 48.6 & 24.0 & 55.0 & 27.7 & 87.0 & {51.7} \\
Gemma-3-4B-it            & 47.2 & 76.8 & 45.2 & 21.0 & 49.0 & 20.7 & 74.0 & {47.7} \\
Qwen-2.5-3B-Instruct       & 34.6 & 65.8 & 41.8 & 14.7 & 35.0 & 21.0 & 80.4 & {41.9} \\
MiniCPM-2B-128k          & 28.4 & 61.6 & 30.8 & 3.7  & 22.0 & 3.3  & 46.6 & {28.1} \\
\midrule
\multicolumn{9}{c}{\textit{Fully Open Models}} \\
\midrule
\textbf{Instella-3B-Long-Instruct} & 43.6 & 73.0 & 51.6 & 30.7 & 54.0 & 32.3 & 84.0 & {52.7} \\
\bottomrule
\end{tabular}
\end{table*}

We evaluate the long-context performance on Helmet~\citep{yen2024helmet}, a recent comprehensive long-context evaluation benchmark encompassing diverse categories. Helmet demonstrates more consistent alignment with human judgment. We evaluate three main tasks across seven datasets: multi-value needle-in-a-haystack (NIAH-MV), retrieval augmented generation (Natural Questions~\citep{kwiatkowski2019natural}, TriviaQA~\citep{joshi2017triviaqa}, HotpotQA~\citep{yang2018hotpotqa}), and long-document QA (InfiniteBench MC/QA~\citep{zhang2024bench}, NarrativeQA~\citep{kovcisky2018narrativeqa}). We use substring exact match (SubEM) for the RAG task, recall for NIAH-MV, and exact match for InfiniteBench MC. For InfiniteBench QA and NarrativeQA, which involve open-ended answers, we rely on gpt-4o-mini to evaluate model responses against the ground truth, following the prompt and metric provided by Helmet. As shown in Table~\ref{tab:long_main_results}, Instella-3B-Long-Instruct outperforms open weights models including Phi-3.5-mini-instruct~\citep{abdin2024phi}, Gemma-3-4B-it~\citep{team2025gemma}, Qwen2.5-3B-Instruct~\citep{yang2024qwen2-5}, and MiniCPM-2B-128k~\citep{hu2024minicpm} on most tasks of the Helmet benchmark. Since the context length of Qwen2.5-3B-Instruct is 32K, we also conduct a side-by-side comparison at 8K, 16K, and 32K context lengths, as shown in Table~\ref{tab:long_qwen_comparison}. Instella-3B-Long-Instruct outperforms Qwen2.5-3B-Instruct by 2.8\% on average.

\begin{table*}[htbp]
\centering
\setlength{\tabcolsep}{5pt}
\caption{\textbf{Comparison with Qwen2.5-3B-Instruct at 8K, 16K, 32K context lengths.}}
\label{tab:long_qwen_comparison}
\begin{tabular}{lccccccccccccc}
\toprule
\multirow{2}{*}{\textbf{Model}} & \multicolumn{3}{c}{\textbf{NIAH-MV}} & \multicolumn{3}{c}{\textbf{NQ}} & \multicolumn{3}{c}{\textbf{TriviaQA}} & \multicolumn{3}{c}{\textbf{HotpotQA}} & \multirow{2}{*}{\textbf{Avg.}} \\
\cmidrule(lr){2-4} \cmidrule(lr){5-7} \cmidrule(lr){8-10} \cmidrule(lr){11-13}
 & {8K} & {16K} & {32K} & {8K} & {16K} & {32K} & {8K} & {16K} & {32K} & {8K} & {16K} & {32K} &  \\
\midrule
Qwen2.5-3B-Instruct       & 95 & 94 & 95 & 48 & 42 & 39 & 77 & 78 & 74 & 51 & 50 & 48 & 65.9 \\
\textbf{Instella-3B-Long-Instruct} & 98 & 95 & 87 & 53 & 49 & 46 & 79 & 73 & 75 & 59 & 59 & 51 & 68.7 \\
\bottomrule
\end{tabular}
\end{table*}

We also evaluate the short-context performance as shown in Table~\ref{tab:long_short_results}. We observe performance drops on some short-context benchmarks compared to Instella-3B-Instruct. Interestingly, TruthfulQA remains stable, Crows-Pairs shows a slight improvement, and the reduction in Toxigen (57.02 → 42.34, lower is better) suggests improved toxicity avoidance, together indicating potential gains in responsible AI benchmarks. We hypothesize that these results reflect a trade-off between optimizing for longer context lengths and retaining short-context performance, which may be more pronounced at the 3B parameter scale compared to larger models.
\begin{table*}[htbp]
\centering
\setlength{\tabcolsep}{6pt}
\caption{\textbf{Evaluation of Instella-Long on general benchmarks.}}
\label{tab:long_short_results}
\begin{tabular}{lcccccc}
\toprule
\textbf{Models} & \textbf{MMLU} & \textbf{IFEval} & \textbf{MT-Bench} & \textbf{TruthfulQA} & \textbf{Toxigen (↓)} & \textbf{Crows-Pair} \\
\midrule
\textbf{Instella-3B-Instruct}        & 58.9 & 71.4 & 7.2 & 55.5 & 57.0 & 58.9 \\
\textbf{Instella-3B-Long-Instruct}   & 57.4 & 68.8 & 6.8 & 55.5 & 42.3 & 60.1 \\
\bottomrule
\end{tabular}
\end{table*}

\subsection{Instella-Math}


Following the same evaluation settings as DeepScaleR-1.5B~\citep{deepscaler2025}, we report Pass@1 accuracy over AIME 2024/25~\citep{AIME}, MATH500~\citep{hendrycks2021measuringmathematicalproblemsolving}, AMC~\citep{AMC12}, Mnerva MATH~\citep{lewkowycz2022solving}, OlympiadBench~\citep{he2024olympiadbenchchallengingbenchmarkpromoting}, GSM8k~\citep{gsm}, and GPQA-Diamond~\citep{rein2023gpqagraduatelevelgoogleproofqa}. Table \ref{tab:math_main_result} reports the Pass@1 rate for the above benchmarks, calculated based on 16 responses per question. Instella-Math delivers competitive performance when compared to leading small-scale open-weight models such as Deepseek-R1-Distilled-Qwen-1.5B, Still-3-1.5B, DeepScaleR-1.5B and SmolLM3-3B. In addition to achieving competitive average performance across all benchmarks, Instella-Math demonstrates the effectiveness of our RL training recipe—improving over its supervised finetuned variant (Instella-Math-SFT) by 10.81 points, compared to a 6.22-point improvement seen in DeepScaleR over its base model (Deepseek-R1-Distilled-Qwen-1.5B).

\begin{table*}[htbp]
\scriptsize
\centering
\setlength{\tabcolsep}{4pt}
\caption{\textbf{Evaluation of Instella-Math on Reasoning Benchmarks}}
\label{tab:math_main_result}
\begin{tabular}{lccccccccc}
\toprule
\textbf{Models} & \textbf{AIME 2024} & \textbf{AIME 2025} & \textbf{MATH500} & \textbf{AMC} & \textbf{Minerva} & \textbf{OlympiadBench} & \textbf{GSM8K} & \textbf{GPQA-D} & \textbf{Avg.} \\ 
\midrule
\multicolumn{10}{c}{\textbf{Pass@1}} \\
\midrule
\multicolumn{10}{c}{\textit{Open-Weight Models}} \\ \midrule
Qwen2.5-Math-1.5B & 7.7 & 4.0 & 57.8 & 35.8 & 15.7 & 26.0 & 66.3 & 15.4 & 28.6 \\
DeepSeek-R1-Distill-Qwen-1.5B & 27.5 & 22.5 & 82.6 & 63.5 & 26.5 & 43.0 & 84.1 & 16.5 & 45.8 \\
STILL-3-1.5B-preview & 30.6 & 25.2 & 84.6 & 66.7 & 28.6 & 45.3 & 86.6 & 19.5 & 48.4 \\
DeepScaleR-1.5B-Preview & 40.6 & 30.8 & 87.4 & 73.2 & 30.1 & 49.9 & 87.3 & 16.5 & 52.0 \\ \midrule
\multicolumn{10}{c}{\textit{Fully-Open Models}} \\ \midrule
OLMo-2-1124-7B-Instruct & 1.3 & 0.2 & 32.6 & 12.3 & 10.3 & 8.5 & 80.9 & 11.1 & 19.6 \\
SmolLM3-3B & 52.5 & 35.8 & 90.2 & 78.7 & 31.8 & 55.4 & 92.3 & 44.9 & 60.2 \\
\textbf{Instella-Math SFT} & 20.0 & 19.0 & 77.6 & 53.9 & 18.8 & 43.3 & 88.0 & 23.4 & 43.0 \\
\textbf{Instella-Math RL Stage 1} & 27.9 & 22.5 & 82.2 & 58.8 & 25.1 & 49.2 & 90.9 & 34.2 & 48.8 \\
\textbf{Instella-Math RL Stage 2} & 29.6 & 22.9 & 85.8 & 66.7 & 27.5 & 52.7 & 91.7 & 37.4 & 51.8 \\
\textbf{Instella-Math RL Stage 3} & 35.6 & 27.7 & 86.5 & 69.7 & 27.7 & 53.1 & 92.5 & 37.6 & {53.8} \\
\midrule
\multicolumn{10}{c}{\textbf{Pass@16}} \\
\midrule
\multicolumn{10}{c}{\textit{Open-Weight Models}} \\ \midrule
Qwen2.5-Math-1.5B & 36.7 & 20.0 & 87.6 & 71.1 & 48.5 & 53.8 & 96.0 & 71.7 & 60.7 \\
DeepSeek-R1-Distill-Qwen-1.5B & 73.3 & 46.7 & 95.0 & 89.2 & 54.4 & 63.9 & 97.0 & 46.5 & 70.7 \\
STILL-3-1.5B-preview & 70.0 & 46.7 & 95.8 & 89.2 & 56.6 & 65.2 & 96.7 & 45.5 & 70.7 \\
DeepScaleR-1.5B-Preview & 70.0 & 53.3 & 95.2 & 91.6 & 54.0 & 66.2 & 96.5 & 39.9 & 70.9 \\ \midrule
\multicolumn{10}{c}{\textit{Fully-Open Models}} \\ \midrule
OLMo-2-1124-7B-Instruct & 13.3 & 3.3 & 66.6 & 50.6 & 35.1 & 23.2 & 97.3 & 49.0 & 42.3 \\
SmolLM3-3B & 76.7 & 77.3 & 96.6 & 94.0 & 54.4 & 72.4 & 98.1 & 90.9 & 82.1 \\
\textbf{Instella-Math SFT} & 50.0 & 40.0 & 94.8 & 89.2 & 44.9 & 64.0 & 97.7 & 83.8 & 70.6 \\
\textbf{Instella-Math RL Stage 1} & 53.3 & 43.3 & 94.6 & 88.0 & 51.5 & 68.6 & 97.6 & 90.9 & 73.5 \\
\textbf{Instella-Math RL Stage 2} & 46.7 & 43.3 & 95.6 & 89.2 & 51.1 & 68.3 & 97.7 & 89.4 & 72.7 \\
\textbf{Instella-Math RL Stage 3} & 63.3 & 50.0 & 95.8 & 86.8 & 50.4 & 68.2 & 97.4 & 88.9 & {75.1} \\
\bottomrule
\end{tabular}
\end{table*}

\begin{table*}[htbp]
\centering
\setlength{\tabcolsep}{4pt}
\caption{\textbf{Evaluation of Instella-Math on TTT-Bench}}
\label{tab:math-TTTbench}
\begin{tabular}{lccccc}
\toprule
\textbf{Models} & \textbf{oTTT} & \textbf{dTTT} & \textbf{cTTT} & \multicolumn{1}{l|}{\textbf{sTTT}} & \multicolumn{1}{c}{\textbf{Avg.}} \\ 
\midrule
\multicolumn{6}{c}{\textit{Open-weight models}} \\ 
\midrule
Qwen2.5-Math-1.5B & 12.5 & 10.0 & 18.9 & \multicolumn{1}{c|}{7.5}  & 12.2 \\
DeepSeek-R1-Distill-Qwen-1.5B & 22.9 & 10.1 & 18.2 & \multicolumn{1}{c|}{3.5} & 13.7 \\
STILL-3-1.5B-preview & 24.5 & 12.3 & 19.8 & \multicolumn{1}{c|}{3.2} & 14.9 \\
DeepScaleR-1.5B-Preview & 23.0 & 16.5 & 23.0 & \multicolumn{1}{c|}{8.2} & 17.7 \\ 
\midrule
\multicolumn{6}{c}{\textit{Fully-open models}} \\ 
\midrule
SmolLM3-3B & 51.2 & 40.1 & 41.3 & \multicolumn{1}{c|}{42.3} & 43.7 \\
\textbf{Instella-Math RL Stage 1} & 56.3 & 31.4 & 39.7 & \multicolumn{1}{c|}{41.9} & 42.3 \\ 
\textbf{Instella-Math RL Stage 2} & 66.2 & 37.3 & 39.2 & \multicolumn{1}{c|}{44.5} & 46.8 \\
\textbf{Instella-Math RL Stage 3} & 70.3 & 39.6 & 40.3 & \multicolumn{1}{c|}{49.0} & 49.8 \\ 
\bottomrule
\end{tabular}
\end{table*}

Additionally, we test Instella-Math on TTT-Bench~\citep{mishra2025tttbenchbenchmarkevaluatingreasoning}, a new benchmark targeting strategic, spatial, and logical reasoning. Remarkably, without any exposure to TTT-Bench–style or similar strategic gaming data during any stage of training, Instella-Math achieves the best performance among all evaluated models (as shown in Table \ref{tab:math-TTTbench}).

More importantly, like OLMo2 and SmolLM-3B, Instella-Math is a fully-open language model, with fully-open training data for the base model (Instella-3B), reasoning SFT, and reinforcement learning stages. In contrast, many competing models are only open-weight releases; their base model training (e.g., Qwen-1.5B) and reasoning distillation processes (e.g., DeepSeek-R1) remain closed.

\section{Conclusion}
 We present Instella, a family of fully open three billion parameter language models that are trained entirely on openly available data and codebase. The Instella model family consists of a strong base pre-trained model, a supervised finetuned instruct model, an 128k token context length long-context model, and a reasoning-centric model. Powered by AMD Instinct™ MI300X GPUs, Instella models attain state-of-the-art performance among fully open models of similar scale and remains competitive with leading open-weight systems despite using notably fewer pre-training tokens. Instella-Long demonstrates strong long-context capabilities, and Instella-Math delivers impressive gains on mathematical and strategic reasoning benchmarks. Alongside model weights, we release the training code, data recipes, and evaluation protocols to support complete reproducibility and transparent benchmarking to foster open-source innovation. Instella models offers a transparent, performant, and extensible foundation for research and application, supporting the community in building more capable and reproducible language models.



\section*{Acknowledgments}
We would like to thank Muthukumar Kumaraswamy, Tanveer Madan Marate, and the AMD IT team for their assistance with the training infrastructure. We also extend our gratitude to Zhenyu Gu, Peng Sun, Steve Reinhardt, Dong Li, and Vikram Appia for their valuable suggestions and support during the model training. Finally, we appreciate the contributions of Cynthia Zamorski, Aarti Choudhary, Hema Chamraj, Ramu Koppineedi, Anshul Gupta, Lindsey Brown and Guruprasad MP in facilitating the model release process.
\bibliography{main}
\bibliographystyle{colm2024_conference}

\end{document}